\documentclass{article}

\usepackage{microtype}
\usepackage{graphicx}
\usepackage{booktabs} 

\usepackage{hyperref}

\usepackage[accepted]{icml2021}

\usepackage{amsmath}
\usepackage{float}
\usepackage{multirow}
\usepackage{wasysym}
\usepackage{xcolor}

\usepackage{tikz}
\usetikzlibrary{bayesnet}

\newcommand{\CRFXO}{CRF$^{\otimes}$}
\newcommand{\CRFX}{CRF$^{\times}$}
\newcommand{\CRFO}{CRF$^{\ocircle}$}

\newcommand{\XX}{$\times$}
\newcommand{\OO}{$\ocircle$}

\newcommand{\A}{\mathbf{A}}
\newcommand{\B}{\mathbf{B}}

\newcommand{\g}{\mathbf{g}}
\newcommand{\h}{\mathbf{h}}
\newcommand{\ii}{\mathbf{i}}
\newcommand{\x}{\mathbf{x}}
\newcommand{\y}{\mathbf{y}}

\newcommand{\YY}{\mathcal{Y}}

\newcommand{\psitth}{\psi(y_{t-1},y_{t};\theta)}
\newcommand{\psitpth}{\psi(y_{t},y_{t+1};\theta)}
\newcommand{\pitth}{\pi(y_{t},\h_{t-2};\theta)}
\newcommand{\phitth}{\phi(y_{t},\h_{t-1};\theta)}
\newcommand{\etatth}{\eta(y_{t},\h_{t};\theta)}
\newcommand{\xitth}{\xi(y_{t},\h_{t+1};\theta)}
\newcommand{\zetatth}{\zeta(y_{t},\h_{t+2};\theta)}
\newcommand{\sigmatth}{\sigma(y_{t},\h_{t-1},\h_{t},\h_{t+1};\theta)}

\newcommand{\psit}{\psi_{t}}
\newcommand{\pit}{\pi_{t}}
\newcommand{\phit}{\phi_{t}}

\newcommand{\etat}{\eta_{t}}
\newcommand{\xit}{\xi_{t}}
\newcommand{\xitm}{\xi_{t-1}}
\newcommand{\zetat}{\zeta_{t}}
\newcommand{\sigmat}{\sigma_{t}}

\newcommand{\f}{\mathbf{f}}
\newcommand{\fpi}{\f^{\pi}}
\newcommand{\fphi}{\f^{\phi}}
\newcommand{\feta}{\f^{\eta}}
\newcommand{\fxi}{\f^{\xi}}
\newcommand{\fzeta}{\f^{\zeta}}
\newcommand{\fsigma}{\f^{\sigma}}

\newcommand{\ayo}{\alpha(y_{1})}
\newcommand{\ayt}{\alpha(y_{t})}
\newcommand{\ayT}{\alpha(y_{T})}
\newcommand{\aytm}{\alpha(y_{t-1})}

\DeclareMathOperator*{\argmax}{arg\,max}

\icmltitlerunning{Locally-Contextual Nonlinear CRFs for Sequence Labeling}

\begin{document}

\twocolumn[
\icmltitle{Locally-Contextual Nonlinear CRFs for Sequence Labeling}



\icmlsetsymbol{equal}{*}

\begin{icmlauthorlist}
\icmlauthor{Harshil Shah}{ucl}
\icmlauthor{Tim Z. Xiao}{ucl}
\icmlauthor{David Barber}{ucl,ati}
\end{icmlauthorlist}

\icmlaffiliation{ucl}{Department of Computer Science, University College London}
\icmlaffiliation{ati}{Alan Turing Institute}


\icmlkeywords{Machine Learning, ICML}

\vskip 0.3in
]



\printAffiliationsAndNotice{}  

\begin{abstract}
    Linear chain conditional random fields (CRFs) combined with contextual word embeddings have achieved state of the art performance on sequence labeling tasks. In many of these tasks, the identity of the neighboring words is often the most useful contextual information when predicting the label of a given word. However, contextual embeddings are usually trained in a task-agnostic manner. This means that although they may encode information about the neighboring words, it is not guaranteed. It can therefore be beneficial to design the sequence labeling architecture to directly extract this information from the embeddings. We propose locally-contextual nonlinear CRFs for sequence labeling. Our approach directly incorporates information from the neighboring embeddings when predicting the label for a given word, and parametrizes the potential functions using deep neural networks. Our model serves as a drop-in replacement for the linear chain CRF, consistently outperforming it in our ablation study. On a variety of tasks, our results are competitive with those of the best published methods. In particular, we outperform the previous state of the art on chunking on CoNLL 2000 and named entity recognition on OntoNotes 5.0 English.
\end{abstract}

\section{Introduction}

In natural language processing, sequence labeling tasks involve labeling every word in a sequence of text with a linguistic tag. These tasks were traditionally performed using shallow linear models such as hidden Markov models (HMMs) \citep{HMMPOS_Kupiec_1992, HMMNER_Bikel_1999} and linear chain conditional random fields (CRFs) \citep{CRF_Lafferty_2001, CRF_McCallum_2003, CRF_Sha_2003}. These approaches model the dependencies between adjacent word-level labels. However when predicting the label for a given word, they do not directly incorporate information from the surrounding words in the sentence (known as `context'). As a result, linear chain CRFs combined with deeper models which do use such contextual information (e.g. convolutional and recurrent networks) gained popularity \citep{ConvCRF_Collobert_2011,SeqLab_Graves_2012,BiLSTMCRF_Huang_2015,BiLSTMCNNCRF_Ma_2016}.

Recently, contextual word embeddings such as those provided by pre-trained language models have become more prominent \citep{ELMo_Peters_2018, GPT_Radford_2018, Flair_Akbik_2018, BERT_Devlin_2019}. Contextual embeddings incorporate sentence-level information, therefore they can be used directly with linear chain CRFs to achieve state of the art performance on sequence labeling tasks \citep{LUKE_Yamada_2020,Dice_Li_2020}.

Contextual word embeddings are typically trained using a generic language modeling objective. This means that the embeddings encode contextual information which can generally be useful for a variety of tasks. However, because these embeddings are not trained for any specific downstream task, there is no guarantee that they will encode the most useful information for that task. Certain tasks such as sentiment analysis and textual entailment typically require global, semantic information about a sentence. In contrast, for sequence labeling tasks it is often the neighboring words in a sentence which are most informative when predicting the label for a given word. Although the contextual embedding of a given word may encode information about its neighboring words, this is not guaranteed. It can therefore be beneficial to design the sequence labeling architecture to directly extract this information from the embeddings \citep{BERTOrNot_Bhattacharjee_2020}.

We therefore propose locally-contextual nonlinear CRFs for sequence labeling. Our approach extends the linear chain CRF in two straightforward ways. Firstly, we directly incorporate information from the neighboring embeddings when predicting the label for a given word. This means that we no longer rely on each contextual embedding to have encoded information about the neighboring words in the sentence. Secondly, we replace the linear potential functions with deep neural networks, resulting in greater modeling flexibility. Locally-contextual nonlinear CRFs can serve as a drop-in replacement for linear chain CRFs, and they have the same computational complexity for training and inference.

We evaluate our approach on chunking, part-of-speech tagging and named entity recognition. On all tasks, our results are competitive with those of the best published methods. In particular, we outperform the previous state of the art on chunking on CoNLL 2000 and named entity recognition on OntoNotes 5.0 English. We also perform an ablation study which shows that both the local context and nonlinear potentials consistently provide improvements compared to linear chain CRFs.

\section{Linear chain CRFs} \label{sec:crf}

\begin{figure}[t]
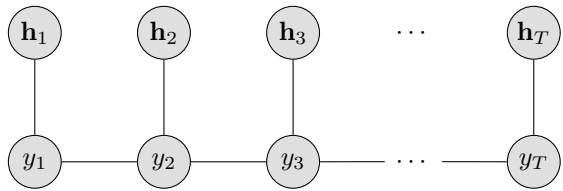

    \centering
    \tikz{%
        \node[obs] (h1) {$\h_{1}$};
        \node[obs, right=of h1] (h2) {$\h_{2}$};
        \node[obs, right=of h2] (h3) {$\h_{3}$};
        \node[const, right=of h3] (hdots) {$\cdots$};
        \node[obs, right=of hdots] (hT) {$\h_{T}$};
        \node[obs, below=of h1] (y1) {$y_{1}$};
        \node[obs, below=of h2] (y2) {$y_{2}$};
        \node[obs, below=of h3] (y3) {$y_{3}$};
        \node[const, below=of hdots, yshift=-15.5pt] (ydots) {$\cdots$};
        \node[obs, below=of hT] (yT) {$y_{T}$};
        \edge[-] {h1} {y1};
        \edge[-] {h2} {y2};
        \edge[-] {h3} {y3};
        \edge[-] {hT} {yT};
        \edge[-] {y1} {y2};
        \edge[-] {y2} {y3};
        \edge[-, shorten >= 4pt] {y3} {ydots};
        \edge[-, shorten <= 4pt] {ydots} {yT};
    }
    \caption{The graphical model of the linear chain CRF.}
    \label{fig:crf}
\end{figure}

Linear chain CRFs \citep{CRF_Lafferty_2001} are popular for various sequence labeling tasks. These include chunking, part-of-speech tagging and named entity recognition, all of which involve labeling every word in a sequence according to a predefined set of labels.

We denote the sequence of words $\x = x_{1}, \hdots, x_{T}$ and the corresponding sequence of labels $\y = y_{1}, \hdots, y_{T}$. We assume that the words have been embedded using either a non-contextual or contextual embedding model; we denote the sequence of embeddings $\h = \h_{1}, \hdots, \h_{T}$. If non-contextual embeddings are used, each embedding is a function only of the word at that time step, i.e. $\h_{t} = \f^{\mathrm{emb-NC}}(x_{t})$. If contextual embeddings are used, each embedding is a function of the entire sentence, i.e. $\h_{t} = \f^{\mathrm{emb-C}}_{t}(\x)$.

The linear chain CRF is shown graphically in Figure \ref{fig:crf}; `linear' here refers to the graphical structure. The conditional distribution of the sequence of labels $\y$ given the sequence of words $\x$ is parametrized as: \begin{align}
    p(\y|\x,\theta) = \frac{\prod_{t} \psit \etat}{\sum_{\y} \prod_{t} \psit \etat}
\end{align} %
where $\theta$ denotes the set of model parameters. The potentials $\psit$ and $\etat$ are defined as: \begin{align}
    \psit &= \psitth \label{eq:crf:def_psi} \\
    \etat &= \etatth \label{eq:crf:def_eta}
\end{align}

The potentials are constrained to be positive and so are parametrized in $\log$-space. Linear chain CRFs typically use $\log$-linear potential functions: \begin{align}
    \log \psit &= \ii(y_{t-1})^{\mathsf{T}} \A \cdot \ii(y_{t}) \label{eq:crf:psi} \\
    \log \etat &= \ii(y_{t})^{\mathsf{T}} \B \cdot \h_{t} \label{eq:crf:eta}
\end{align}%
where $\ii(y)$ denotes the one-ht encoding of $y$, and $\A$ and $\B$ are parameters of the model. Note that `$\log$-linear' here refers to linearity in the parameters. Henceforth, we refer to linear chain CRFs simply as CRFs.

\subsection{Incorporating contextual information} \label{sec:crf:ctx}

When predicting the label at a given time step $y_{t}$, it is often necessary to use information from words in the sentence other than only $x_{t}$; we refer to this as contextual information. For example, below are two sentences from the CoNLL 2003 named entity recognition training set \citep{CoNLL_Tjong_2003}:

\begin{table}[H]
    \centering
    \begin{tabular}{p{0.9\linewidth}}
        \toprule
        {One had threatened to blow it up unless it was refuelled and they were taken to [\textbf{\textit{LOC}} London] where they intended to surrender and seek political asylum.} \\
        \midrule
        {[\textbf{\textit{ORG}} St Helens] have now set their sights on taking the treble by winning the end-of-season premiership which begins with next Sunday's semifinal against [\textbf{\textit{ORG}} London].} \\
        \bottomrule
    \end{tabular}
\end{table}

In each example, the word ``London'' ($x_{t}$) has a different label ($y_{t}$) and so it is necessary to use contextual information to make the correct prediction.

In Figure \ref{fig:crf}, we see that there is no direct path from any $\h_{j}$ with $j \neq t$ to $y_{t}$. This means that in order to use contextual information, CRFs rely on either/both of the following: \begin{itemize}
    \item The transition potentials $\psitth$ and $\psitpth$ carrying this information from the labels at positions $t - 1$ and $t + 1$.
    \item The (contextual) embedding $\h_{t}$ encoding information about words $x_{j}$ with $j \neq t$.
\end{itemize}

Relying on the transition potentials may not always be effective because knowing the previous/next label is often not sufficient when labeling a given word. In the above examples, the previous/next labels indicate that they are not part of a named entity (i.e. $y_{t-1} = \mathrm{O}$, $y_{t+1} = \mathrm{O}$). Knowing this does not help to identify whether ``London'' refers to a location, an organization, or even a person's name.

In the first example, knowing that the previous word ($x_{t-1}$) is ``to'' and the next word ($x_{t+1}$) is ``where'' indicates that ``London'' is a location. In the second example, knowing that the previous word ($x_{t-1}$) is ``against'' indicates that ``London'' is a sports organization. Therefore, to label these sentences correctly, a CRF relies on the embedding $\h_{t}$ to encode the identities of the neighboring words.

However, contextual embedding models are usually trained in a manner that is agnostic to the downstream tasks they will be used for. Tasks such as sentiment analysis and textual entailment typically require global sentence-level context whereas sequence labeling tasks usually require local contextual information from the neighboring words in the sentence, as shown in the example above. Because different downstream tasks require different types of contextual information, it is not guaranteed that task-agnostic contextual embeddings will encode the most useful contextual information for any specific task. It can therefore be beneficial to design the architecture for the downstream task to directly extract the most useful information from the embeddings.

\section{The \CRFXO{} model} \label{sec:model}

\begin{figure}[t]
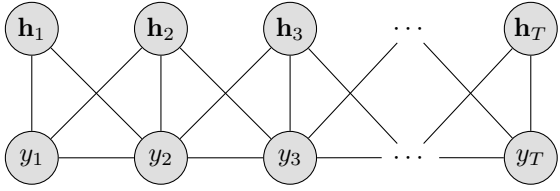

    \centering
    \tikz{%
        \node[obs] (h1) {$\h_{1}$};
        \node[obs, right=of h1] (h2) {$\h_{2}$};
        \node[obs, right=of h2] (h3) {$\h_{3}$};
        \node[const, right=of h3] (hdots) {$\cdots$};
        \node[obs, right=of hdots] (hT) {$\h_{T}$};
        \node[obs, below=of h1] (y1) {$y_{1}$};
        \node[obs, below=of h2] (y2) {$y_{2}$};
        \node[obs, below=of h3] (y3) {$y_{3}$};
        \node[const, below=of hdots, yshift=-15.5pt] (ydots) {$\cdots$};
        \node[obs, below=of hT] (yT) {$y_{T}$};
        \edge[-] {h1} {y1};
        \edge[-] {h2} {y2};
        \edge[-] {h3} {y3};
        \edge[-] {hT} {yT};
        \edge[-] {h1} {y2};
        \edge[-] {h2} {y3};
        \edge[-, shorten >= 4pt] {h3} {ydots};
        \edge[-, shorten <= 4pt] {hdots} {yT};
        \edge[-] {y1} {h2};
        \edge[-] {y2} {h3};
        \edge[-, shorten >= 4pt] {y3} {hdots};
        \edge[-, shorten <= 4pt] {ydots} {hT};
        \edge[-] {y1} {y2};
        \edge[-] {y2} {y3};
        \edge[-, shorten >= 4pt] {y3} {ydots};
        \edge[-, shorten <= 4pt] {ydots} {yT};
    }
    \caption{The graphical model of the \CRFXO{}.}
    \label{fig:model}
\end{figure}

We introduce the \CRFXO{}, which extends the CRF by directly using the neighboring embeddings when predicting the label for a given word. We also replace the $\log$-linear potential functions with deep neural networks to provide greater modeling flexibility. We call our model \CRFXO{}, where \XX{} refers to the locally-contextual structure and \OO{} refers to the nonlinear potentials.

The graphical model is shown in Figure \ref{fig:model}. The conditional distribution of the labels $\y$ given the sentence $\x$ is parametrized as follows: \begin{align}
    p(\y|\x,\theta) = \frac{\prod_{t} \psit \phit \etat \xit}{\sum_{\y} \prod_{t} \psit \phit \etat \xit}
\end{align} %
where the additional potentials $\phit$ and $\xit$ are defined as: \begin{align}
    \phit &= \phitth \label{eq:model:def_phi} \\
    \xit &= \xitth \label{eq:model:def_xi}
\end{align}

With this structure, the embeddings $\h_{t-1}$ and $\h_{t+1}$, in addition to $\h_{t}$, are directly used when modeling the label $y_{t}$. This means that the model no longer relies on the embedding $\h_{t}$ to have encoded information about the neighboring words in the sentence. We evaluate a more general parametrization in Appendix \ref{sec:apx:alt:param} but find its empirical performance to be worse. We also evaluate a wider contextual window in Appendix \ref{sec:apx:alt:wide} but find that it does not provide improved results.

Since the labels $y_{t}$ are discrete, the parametrization of $\log \psit$ remains the same as in Equation (\ref{eq:crf:psi}). However unlike the $\log$-linear parametrization in Equation (\ref{eq:crf:eta}), $\log \etat$ along with $\log \phit$ and $\log \xit$ are each parametrized by feedforward networks which take as input the embeddings $\h_{t-1}$, $\h_{t}$ and $\h_{t+1}$ respectively \citep{CNF_Peng_2009}: \begin{align}
    & \log \phit = \ii(y_{t})^{\mathsf{T}} \fphi(\h_{t-1}) \label{eq:model:phi} \\
    & \log \etat = \ii(y_{t})^{\mathsf{T}} \feta(\h_{t}) \label{eq:model:eta} \\
    & \log \xit = \ii(y_{t})^{\mathsf{T}} \fxi(\h_{t+1}) \label{eq:model:xi}
\end{align}
where $\fphi$, $\feta$ and $\fxi$ are the feedforward networks.

\begin{figure}[t]
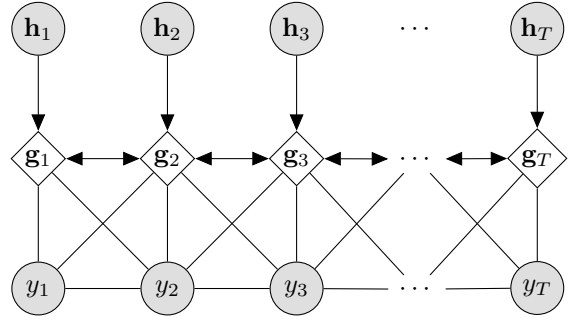

    \centering
    \tikz{%
        \node[obs] (h1) {$\h_{1}$};
        \node[obs, right=of h1] (h2) {$\h_{2}$};
        \node[obs, right=of h2] (h3) {$\h_{3}$};
        \node[const, right=of h3] (hdots) {$\cdots$};
        \node[obs, right=of hdots] (hT) {$\h_{T}$};
        \node[det, below=of h1] (g1) {$\g_{1}$};
        \node[det, below=of h2] (g2) {$\g_{2}$};
        \node[det, below=of h3] (g3) {$\g_{3}$};
        \node[const, below=of hdots, yshift=-16pt] (gdots) {$\cdots$};
        \node[det, below=of hT, yshift=1pt] (gT) {$\g_{T}$};
        \node[obs, below=of g1] (y1) {$y_{1}$};
        \node[obs, below=of g2] (y2) {$y_{2}$};
        \node[obs, below=of g3] (y3) {$y_{3}$};
        \node[const, below=of gdots, yshift=-16pt] (ydots) {$\cdots$};
        \node[obs, below=of gT, yshift=1pt] (yT) {$y_{T}$};
        \edge {h1} {g1};
        \edge {h2} {g2};
        \edge {h3} {g3};
        \edge {hT} {gT};
        \edge[<->] {g1} {g2};
        \edge[<->] {g2} {g3};
        \edge[<->, shorten >= 4pt] {g3} {gdots};
        \edge[<->, shorten <= 4pt] {gdots} {gT};
        \edge[-] {g1} {y1};
        \edge[-] {g2} {y2};
        \edge[-] {g3} {y3};
        \edge[-] {gT} {yT};
        \edge[-] {g1} {y2};
        \edge[-] {g2} {y3};
        \edge[-, shorten >= 4pt] {g3} {ydots};
        \edge[-, shorten <= 4pt] {gdots} {yT};
        \edge[-] {y1} {g2};
        \edge[-] {y2} {g3};
        \edge[-, shorten >= 4pt] {y3} {gdots};
        \edge[-, shorten <= 4pt] {ydots} {gT};
        \edge[-] {y1} {y2};
        \edge[-] {y2} {y3};
        \edge[-, shorten >= 4pt] {y3} {ydots};
        \edge[-, shorten <= 4pt] {ydots} {yT};
        
    }
    \caption{The graphical model of the \CRFXO{} combined with a bidirectional LSTM.}
    \label{fig:model:bilstm}
\end{figure}

\subsection*{Training}

We train the model to maximize $\log p(\y|\x,\theta)$ with respect to the set of parameters $\theta = \{\A, \phi, \eta, \xi\}$. The objective can be expressed as: \begin{align}
    \log p(\y|\x,\theta) = & \sum_{t} \left[ \log \psit + \log \phit + \log \etat + \log \xit \right] \nonumber \\
    & - \log \sum_{\y} \prod_{t} \psit \phit \etat \xit
\end{align}

The first term is straightforward to compute. The second term can be computed using dynamic programming, analogous to computing the likelihood in HMMs \citep{HMM_Rabiner_1989}. We initialize: \begin{align}
    \ayo = \psi_{1} \eta_{1}
\end{align}

Then, for $t = 2, \hdots, T$: \begin{align}
    \ayt = \sum_{y_{t-1}} \aytm \psit \phit \etat \xitm \label{eq:model:train:rec}
\end{align}

Finally: \begin{align}
    \sum_{\y} \prod_{t} \psit \phit \etat \xit = \sum_{y_{T}} \ayT \label{eq:model:train:fin}
\end{align}

Denoting $\YY$ as the set of possible labels, the time complexity of this recursion is $O(T|\YY|^{2})$, the same as for the CRF.

\subsection*{Inference}

During inference, we want to find $\y^{*}$ such that: \begin{align}
    \y^{*} = \argmax_{\y} \ \log p(\y|\x)
\end{align}

This can be done using the Viterbi algorithm \citep{Viterbi_Viterbi_1967}, which simply replaces the sum in Equation (\ref{eq:model:train:rec}) with a $\max$ operation.

The key advantage of the \CRFXO{} is that it uses enhanced local contextual information while retaining the computational tractability and parsimony of the CRF. As discussed in Section \ref{sec:crf:ctx}, sequence labeling tasks will benefit from this enhanced local context, as demonstrated empirically in the experiments and ablation study in Sections \ref{sec:exp} and \ref{sec:abl} respectively.

\subsection{Combination with bi-LSTMs} \label{sec:model:bilstm}

\citet{BiLSTMCRF_Huang_2015} augment the CRF with a bidirectional LSTM layer in order to use global, sentence-level contextual information when predicting the label for a given word. The \CRFXO{} can similarly be combined with a bidirectional LSTM layer.

Figure \ref{fig:model} shows that with the \CRFXO{}, there is still no direct path from any $\h_{t \pm j}$ with $j \geq 2$ to $y_{t}$. Therefore, particularly when using non-contextual embeddings, encoding wider contextual information by using a bidirectional LSTM can be very useful.


The graphical model of the \CRFXO{} combined with a bidirectional LSTM is shown in Figure \ref{fig:model:bilstm}. The sequence of embeddings $\h_{1}, \hdots, \h_{T}$ is fed to a bidirectional LSTM to produce a sequence of states $\g_{1}, \hdots, \g_{T}$. These states are then used as the inputs to the feedforward networks in Equations (\ref{eq:model:phi}) to (\ref{eq:model:xi}) instead of the embeddings.

\section{Related work} \label{sec:rel}

There have been several approaches to directly incorporate contextual information into sequence labeling architectures. The Conv-CRF \citep{ConvCRF_Collobert_2011}, biLSTM-CRF \citep{BiLSTMCRF_Huang_2015}, and biLSTM-CNN-CRF \citep{BiLSTMCNNCRF_Ma_2016} each augment the CRF with either a convolutional network, a bidirectional LSTM, or both in order to use contextual features when labeling a given word. More recently, \citet{SentenceStateLSTM_Zhang_2018} propose an alternative LSTM structure for encoding text which consists of a parallel state for each word, achieving improved results over the biLSTM-CRF for sequence labeling tasks. GCDT \citep{GCDT_Liu_2019} improves an RNN-based architecture by augmenting it with a sentence-level representation which captures wider contextual information. \citet{HierBiLSTM_Luo_2020} learn representations encoding both sentence-level and document-level features for named entity recognition using a hierarchical bidirectional LSTM. \citet{XSentContexts_Luoma_2020} leverage the fact that BERT can represent inputs consisting of several sentences in order to use cross-sentence context when performing named entity recognition.

In work done concurrently to ours, \citet{CRFXO_Hu_2020} also evaluate locally-contextual parametrisations of the CRF and find that the local context consistently improves results.

We compare our approach against the best of these methods in Section \ref{sec:exp}.

\section{Datasets} \label{sec:data}

\begin{table*}[t]
    \centering
    \begin{tabular}{llllll}
        \toprule
        \textsc{Dataset} & \textsc{Task} & \textsc{Labels} & \textsc{Train} & \textsc{Validation} & \textsc{Test} \\
        \midrule
        CoNLL 2000 & Chunking & 11 & 7,936* & 1,000* & 2,012 \\
        Penn Treebank & POS & 45 & 38,219 & 5,527 & 5,462 \\
        CoNLL 2003 & NER & 4 & 14,987 & 3,466 & 3,684 \\
        OntoNotes & NER & 18 & 59,924 & 8,528 & 8,262 \\
        \bottomrule
    \end{tabular}
    \caption{Statistics of each of the datasets used. We use the standard splits for all datasets.}
    \vskip 2pt
    \footnotesize{*The CoNLL 2000 dataset does not include a validation set. We therefore randomly sample 1,000 sentences from the training set to use for validation.}
    \label{tab:data:stats}
\end{table*}

\begin{table}[t]
    \centering
    \begin{tabular}{lc}
        \toprule
        \textsc{Model} & \textsc{F1} \\
        \midrule
        \CRFXO{}(GloVe)          & 96.12 \\
        \CRFXO{}(GloVe, biLSTM)  & 96.14 \\
        \CRFXO{}(BERT)           & 97.40 \\
        \CRFXO{}(Flair)          & \textbf{97.52} \\
        \midrule
        \citet{GCDT_Liu_2019}    & 97.30 \\
        \citet{CVT_Clark_2018}   & 97.00 \\
        \citet{Flair_Akbik_2018} & 96.72 \\
        \bottomrule
    \end{tabular}
    \caption{Chunking results on the test set of the CoNLL 2000 dataset.}
    \label{tab:exp:res:chunk}
    \vskip 8pt
\end{table}

\begin{table}[t]
    \centering
    \begin{tabular}{lc}
        \toprule
        \textsc{Model} & \textsc{Accuracy} \\
        \midrule
        \CRFXO{}(GloVe)                & 97.15 \\
        \CRFXO{}(GloVe, biLSTM)        & 97.15 \\
        \CRFXO{}(BERT)                 & 97.24 \\
        \CRFXO{}(Flair)                & 97.56 \\
        \midrule
        \citet{MetaBiLSTM_Bohnet_2018} & \textbf{97.96} \\
        \citet{Flair_Akbik_2018}       & 97.85 \\
        \citet{CharBiLSTM_Ling_2015}   & 97.78 \\
        \bottomrule
    \end{tabular}
    \caption{Part-of-speech tagging results on the test set of the Penn Treebank dataset.}
    \label{tab:exp:res:ptb}
\end{table}

\begin{table}[t]
    \centering
    \begin{tabular}{lc}
        \toprule
        \textsc{Model} & \textsc{F1} \\
        \midrule
        \CRFXO{}(GloVe)                  & 91.81 \\
        \CRFXO{}(GloVe, biLSTM)          & 91.34 \\
        \CRFXO{}(BERT)                   & 93.82 \\
        \CRFXO{}(Flair)                  & 94.22 \\
        \midrule
        \citet{LUKE_Yamada_2020}         & \textbf{94.30} \\
        \citet{XSentContexts_Luoma_2020} & 93.74 \\
        \citet{Cloze_Baevski_2019}       & 93.50 \\
        \bottomrule
    \end{tabular}
    \caption{Named entity recognition results on the test set of the CoNLL 2003 dataset.}
    \label{tab:exp:res:conll}
    \vskip 8pt
\end{table}

\begin{table}[t]
    \centering
    \begin{tabular}{lc}
        \toprule
        \textsc{Model} & \textsc{F1} \\
        \midrule
        \CRFXO{}(GloVe)         & 88.70 \\
        \CRFXO{}(GloVe, biLSTM) & 89.04 \\
        \CRFXO{}(BERT)          & \textbf{92.17} \\
        \CRFXO{}(Flair)         & 91.54 \\
        \midrule
        \citet{Dice_Li_2020}    & 92.07 \\
        \citet{NERDep_Yu_2020}  & 91.30 \\
        \citet{MRC_Li_2020}     & 91.11 \\
        \bottomrule
    \end{tabular}
    \caption{Named entity recognition results on the test set of the OntoNotes 5.0 dataset.}
    \label{tab:exp:res:onto}
\end{table}

\begin{figure*}[t]
    \centering
    \includegraphics[width=0.48\linewidth]{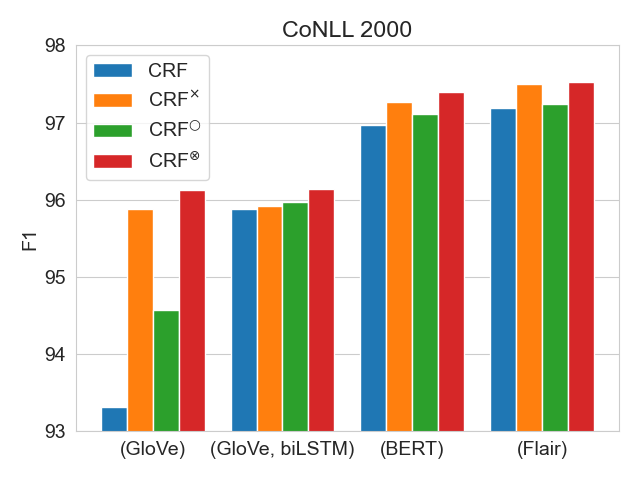}
    \hfill
    \includegraphics[width=0.48\linewidth]{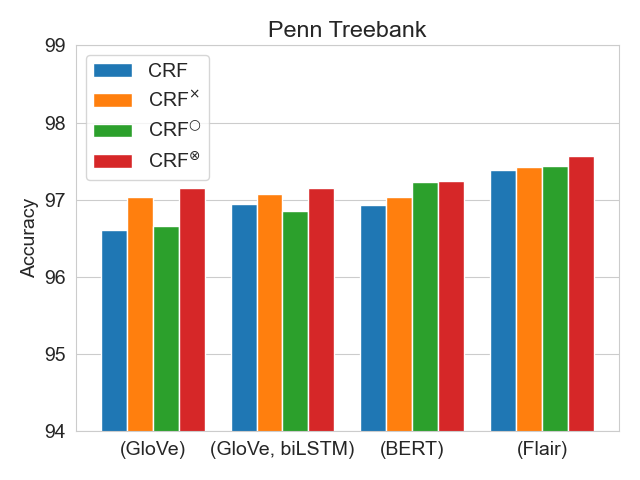}
    
    \vskip 6pt
    
    \includegraphics[width=0.48\linewidth]{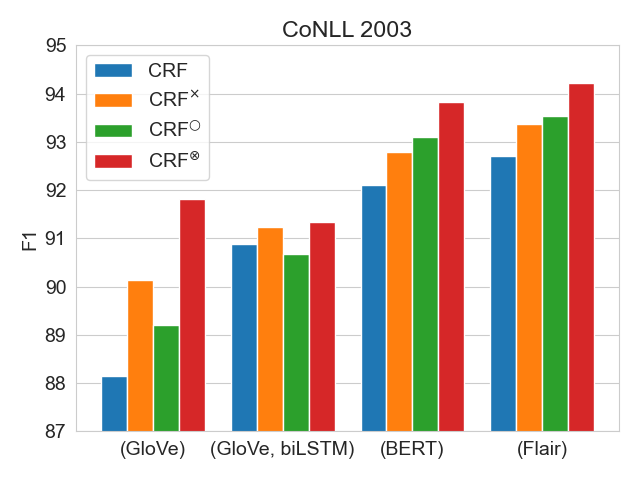}
    \hfill
    \includegraphics[width=0.48\linewidth]{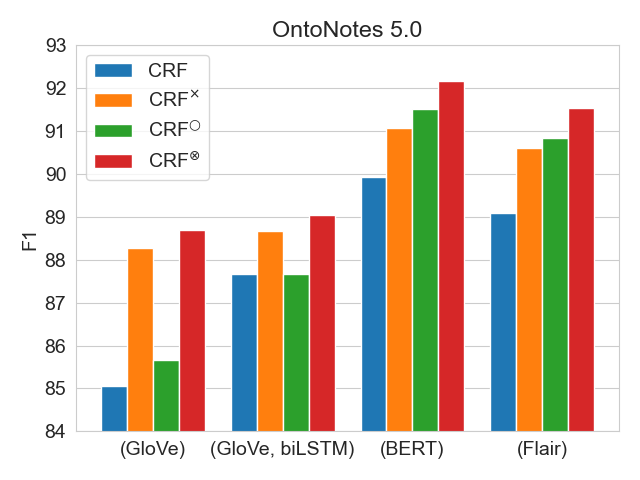}
    
    \vskip -6pt
    
    \caption{Results of the ablation study on each of the test sets. We compare the results when augmenting the CRF with locally-contextual connections, nonlinear potential functions, and both.}
    \label{fig:abl:res}
\end{figure*}

We evaluate the \CRFXO{} on the classic sequence labeling tasks of chunking, part-of-speech tagging, and named entity recognition. We use the following datasets (summary statistics for each are shown in Table \ref{tab:data:stats}):

\paragraph{Chunking}

Chunking consists of dividing sentences into syntactically correlated parts of words according to a set of predefined chunk types. This task is evaluated using the span F1 score.

We use the CoNLL 2000 dataset \citep{CoNLL_Tjong_2000}. An example sentence from the training set is shown below:

\begin{table}[H]
    \centering
    \begin{tabular}{p{0.9\linewidth}}
        \toprule
        {[\textbf{\textit{NP}} He] [\textbf{\textit{VP}} reckons] [\textbf{\textit{NP}} the current account deficit] [\textbf{\textit{VP}} will narrow] [\textbf{\textit{PP}} to] [\textbf{\textit{NP}} only \# 1.8 billion] [\textbf{\textit{PP}} in] [\textbf{\textit{NP}} September].} \\
        \bottomrule
    \end{tabular}
\end{table}

\paragraph{Part-of-speech tagging}

Part-of-speech tagging consists of labeling each word in a sentence according to its part-of-speech. This task is evaluated using accuracy.

We use the Wall Street Journal portion of the Penn Treebank dataset \citep{PTB_Marcus_1993}. An example sentence from the training set is shown below:

\begin{table}[H]
    \centering
    \begin{tabular}{p{0.9\linewidth}}
        \toprule
        {[\textbf{\textit{NN}} Compound] [\textbf{\textit{NNS}} yields] [\textbf{\textit{VBP}} assume] [\textbf{\textit{NN}} reinvestment] [\textbf{\textit{IN}} of] [\textbf{\textit{NNS}} dividends] [\textbf{\textit{CC}} and] [\textbf{\textit{IN}} that] [\textbf{\textit{DT}} the] [\textbf{\textit{JJ}} current] [\textbf{\textit{NN}} yield] [\textbf{\textit{VBZ}} continues] [\textbf{\textit{IN}} for] [\textbf{\textit{DT}} a] [\textbf{\textit{NN}} year].} \\
        \bottomrule
    \end{tabular}
\end{table}

\paragraph{Named entity recognition}

Named entity recognition consists of locating and classifying named entities in sentences. The entities are classified under a pre-defined set of entity categories.

We use the CoNLL 2003 English \citep{CoNLL_Tjong_2003} and OntoNotes 5.0 English \citep{OntoNotes_Pradhan_2013} datasets. The CoNLL 2003 dataset consists of 4 high-level entity types while the OntoNotes 5.0 dataset involves 18 fine-grained entity types. An example from the OntoNotes 5.0 training set is shown below:

\begin{table}[H]
    \centering
    \begin{tabular}{p{0.9\linewidth}}
        \toprule
        {[\textbf{\textit{NORP}} Japanese] Prime Minister [\textbf{\textit{PERSON}} Junichiro Koizumi] also arrived in [\textbf{\textit{GPE}} Pusan] [\textbf{\textit{TIME}} this afternoon], beginning his [\textbf{\textit{ORG}} APEC] trip this time.} \\
        \bottomrule
    \end{tabular}
\end{table}

These are all important tasks in natural language processing and play vital roles in downstream tasks such as dependency parsing, question answering and relation extraction. Even small improvements on sequence labeling tasks can provide significant benefits for these downstream tasks \citep{DepPars_Nguyen_2018,QA_Park_2015,Rel_Liu_2017,LUKE_Yamada_2020}.

\section{Experiments} \label{sec:exp}

\begin{table*}[t]
    \centering
    \begin{tabular}{p{1.5cm} p{3cm} p{11cm}}
        \toprule
        \textsc{Dataset} & \textsc{Model} & \textsc{Prediction} \\
        \midrule
        \multirow{11}{1.5cm}{CoNLL 2003} & Ground truth \& \par \CRFXO{}(BERT) \& \par \CRFXO{}(Flair) & {[\textbf{\textit{PER}} Mike Cito], 17, was expelled from [\textbf{\textit{ORG}} St Pius X High School] in [\textbf{\textit{LOC}} Albuquerque] after an October game in which he used the sharpened chin strap buckles to injure two opposing players and the referee.} \\[12pt]
         & CRF(BERT) & {[\textbf{\textit{PER}} Mike Cito], 17, was expelled from [\textbf{\textit{ORG}} St Pius X] High School in [\textbf{\textit{LOC}} Albuquerque] after an October game in which he used the sharpened chin strap buckles to injure two opposing players and the referee.} \\[12pt]
         & CRF(Flair) & {[\textbf{\textit{PER}} Mike Cito], 17, was expelled from [\textbf{\textit{LOC}} St Pius X High School] in [\textbf{\textit{LOC}} Albuquerque] after an October game in which he used the sharpened chin strap buckles to injure two opposing players and the referee.} \\
        \midrule
        \multirow{9}{1.5cm}{OntoNotes 5.0} & Ground truth \& \par \CRFXO{}(BERT) \& \par \CRFXO{}(Flair) & {It is [\textbf{\textit{TIME}} three hours] by car to [\textbf{\textit{GPE}} Hong Kong] and [\textbf{\textit{TIME}} one and a half hours] by boat to [\textbf{\textit{GPE}} Humen].} \\[25pt]
         & CRF(BERT) & {It is [\textbf{\textit{CARDINAL}} three] hours by car to [\textbf{\textit{GPE}} Hong Kong] and [\textbf{\textit{TIME}} one and a half hours] by boat to [\textbf{\textit{GPE}} Humen].} \\[12pt]
         & CRF(Flair) & {It is [\textbf{\textit{TIME}} three hours] by car to [\textbf{\textit{GPE}} Hong Kong] and [\textbf{\textit{CARDINAL}} one] and [\textbf{\textit{TIME}} a half hours] by boat to [\textbf{\textit{GPE}} Humen].} \\
        \bottomrule
    \end{tabular}
    \caption{Example sentences from the named entity recognition test sets where the \CRFXO{}(BERT) and \CRFXO{}(Flair) models make the correct predictions but the CRF{}(BERT) and CRF{}(Flair) models make the incorrect predictions.}
    \label{tab:abl:eg:right}
\end{table*}

\begin{table*}[t]
    \centering
    \begin{tabular}{p{1.5cm} p{3cm} p{11cm}}
        \toprule
        \textsc{Dataset} & \textsc{Model} & \textsc{Prediction} \\
        \midrule
        \multirow{5}{1.5cm}{CoNLL 2003} & Ground truth & {[\textbf{\textit{ORG}} Bre-X], [\textbf{\textit{ORG}} Barrick] said to continue [\textbf{\textit{LOC}} Busang] talks.} \\[12pt]
         & \CRFXO{}(BERT) & {[\textbf{\textit{ORG}} Bre-X], [\textbf{\textit{PER}} Barrick] said to continue [\textbf{\textit{ORG}} Busang] talks.} \\[12pt]
         & \CRFXO{}(Flair) & {[\textbf{\textit{LOC}} Bre-X], [\textbf{\textit{ORG}} Barrick] said to continue [\textbf{\textit{MISC}} Busang] talks.} \\
        \midrule
        \multirow{8}{1.5cm}{OntoNotes 5.0} & Ground truth & {In the near future, [\textbf{\textit{EVENT}} the Russian Tumen River Region Negotiation Conference] will also be held in [\textbf{\textit{GPE}} Vladivostok].} \\[12pt]
         & \CRFXO{}(BERT) & {In the near future, the [\textbf{\textit{NORP}} Russian] [\textbf{\textit{LOC}} Tumen River] [\textbf{\textit{ORG}} Region Negotiation Conference] will also be held in [\textbf{\textit{GPE}} Vladivostok].} \\[12pt]
         & \CRFXO{}(Flair) & {In the near future, the [\textbf{\textit{NORP}} Russian] [\textbf{\textit{LOC}} Tumen River Region] [\textbf{\textit{EVENT}} Negotiation Conference] will also be held in [\textbf{\textit{GPE}} Vladivostok].} \\
        \bottomrule
    \end{tabular}
    \caption{Example sentences from the named entity recognition test sets where the \CRFXO{}(BERT) and \CRFXO{}(Flair) models make the incorrect predictions.}
    \label{tab:abl:eg:wrong}
\end{table*}

\subsection{Model architectures and training} \label{sec:exp:model}

We train four different versions of the \CRFXO{} and compare against the best published results on each dataset. The variants are as follows: \begin{itemize}
    \item \CRFXO{}(GloVe):
    \begin{itemize}
        \item This uses 300-dimensional GloVe embeddings \citep{GloVe_Pennington_2014} combined with the \CRFXO{}.
    \end{itemize}
    
    \item \CRFXO{}(GloVe, biLSTM)
    \begin{itemize}
        \item This uses 300-dimensional GloVe embeddings combined with a biLSTM and the \CRFXO{}. The forward and backward LSTM states each have 300 units.
    \end{itemize}
    
    \item \CRFXO{}(BERT)
    \begin{itemize}
        \item This uses the final (768-dimensional) layer of the BERT-Base model \citep{BERT_Devlin_2019} with the \CRFXO{}. BERT embeddings are defined on sub-word tokens rather than on words. Therefore if the BERT tokenizer splits a word into multiple sub-word tokens, we take the mean of the sub-word embeddings as the word embedding.
    \end{itemize}
    
    \item \CRFXO{}(Flair)
    \begin{itemize}
        \item This uses Flair embeddings \citep{Flair_Akbik_2018} with the \CRFXO{}. \citet{Flair_Akbik_2018} recommend concatenating 100-dimensional GloVe embeddings to their forward and backward language model embeddings. The resulting Flair embeddings are 4196-dimensional.
    \end{itemize}
\end{itemize}

Note that the GloVe embeddings are non-contextual whereas the BERT and Flair embeddings are contextual. Throughout training, we update the embeddings in the GloVe versions but do not update the BERT or Flair embedding models.

In our initial experiments, we also trained \CRFXO{}(BERT, biLSTM) and \CRFXO{}(Flair, biLSTM) models but found that these did not improve performance compared to using the \CRFXO{} without the bidirectional LSTM. This is unsurprising because the BERT and Flair embeddings are already trained to encode sentence-level context.

In each model, for the feedforward networks $\fphi$, $\feta$ and $\fxi$ referred to in Equations (\ref{eq:model:phi}), (\ref{eq:model:eta}) and (\ref{eq:model:xi}), we use 2 layers with 600 units, ReLU activations and a skip connection. We train each model using stochastic gradient descent with a learning rate of 0.001 and Nesterov momentum of 0.9 \citep{NAG_Nesterov_1983}. We train for a maximum of 100,000 iterations, using early stopping on the validation set.

\subsection{Results}

The results for the four datasets are shown in Tables \ref{tab:exp:res:chunk}, \ref{tab:exp:res:ptb}, \ref{tab:exp:res:conll}, and \ref{tab:exp:res:onto}. Across all of the tasks, our results are competitive with the best published methods. We find that the previous state of the art is outperformed by both \CRFXO{}(BERT) and \CRFXO{}(Flair) on chunking on CoNLL 2000 and by \CRFXO{}(BERT) on named entity recognition on OntoNotes 5.0.

We find that on all of the tasks, the BERT and Flair versions outperform both of the GloVe versions. With \CRFXO{}(Glove, biLSTM), the bidirectional LSTM does encode sentence-level contextual information. However, it is trained on a much smaller amount of data than the BERT and Flair embedding models, likely resulting in its lower scores. 

In general, there is little difference in performance between \CRFXO{}(Glove) and \CRFXO{}(Glove, biLSTM). This may suggest that the local context used by the \CRFXO{} is sufficient for these tasks and that the wider context provided by the bidirectional LSTM does not add significant value when using the \CRFXO{}. This is consistent with the results provided in Appendix \ref{sec:apx:alt:wide}.


\section{Ablation study} \label{sec:abl}

In this section, we attempt to understand the effects of the local context and nonlinear potentials when compared to the CRF.

To this end, we train three additional variants of each of the model versions described in Section \ref{sec:exp:model}. In each case, we replace the \CRFXO{} component with one of the following: \begin{itemize}
    \item \CRFX{}
    \begin{itemize}
        \item This is the same as the \CRFXO{}, but with linear (instead of nonlinear) potential functions.
    \end{itemize}
    
    \item \CRFO{}
    \begin{itemize}
        \item This removes the local context, i.e. it is the same as the \CRFXO{} but with the $\phit$ and $\xit$ potentials removed. The $\etat$ potential function is still nonlinear.
    \end{itemize}
    
    \item CRF
    \begin{itemize}
        \item This is the linear chain CRF. It is the same as the \CRFO{}, but the $\etat$ potential function is linear instead of nonlinear.
    \end{itemize}
\end{itemize}

\subsection{Results}

Figure \ref{fig:abl:res} shows the results with each of these variants. In all cases, we see that the local context is helpful: for each set of embeddings, the results for the \CRFXO{} are higher than for the \CRFO{} and the results for the \CRFX{} are higher than for the CRF. The story is similar for the nonlinear potentials; in almost all cases, the \CRFXO{} is better than the \CRFX{} and the \CRFO{} is better than the CRF.

Table \ref{tab:abl:eg:right} shows example sentences from the two named entity recognition test sets where \CRFXO{}(BERT) and \CRFXO{}(Flair) make correct predictions but CRF(BERT) and CRF(Flair) do not. In the first example, CRF(BERT) mistakes the length of the organization ``St Pius X High School''. This does not happen with \CRFXO{}(BERT), most likely because the model is aware that the word after ``High'' is ``School'' and therefore that this is a single organization entity. Similarly, in the second example both CRF(BERT) and CRF(Flair) struggle to identify whether or not the quantities are cardinals or if they are related to the time measurements. In contrast, \CRFXO{}(BERT) and \CRFXO{}(Flair) do not suffer from the same mistakes.

While the \CRFXO{} models work well, it is interesting to examine where they still make errors. Table \ref{tab:abl:eg:wrong} shows example sentences from the named entity recognition test sets where \CRFXO{}(BERT) and \CRFXO{}(Flair) make incorrect predictions. We see that in both examples, both models correctly identify the positions of the named entities but struggle with their types. The first example is one that even a human may struggle to label without any external knowledge beyond the sentence itself. In the second example, both models fail to recognize the one long entity and instead break it down into several smaller, plausible entities.

\subsection{Computational efficiency}

Table \ref{tab:apx:abl:comp} in Appendix \ref{sec:apx:abl:comp} shows the detailed computational efficiency statistics of each of the models we train. In summary, we find that the \CRFXO{} takes approximately 1.5 to 2 times longer for each training and inference iteration than the CRF. The \CRFX{} and \CRFXO{} models have approximately 3 times as many parameters as the CRF and \CRFO{} respectively.

\section{Discussion and future work}

We propose locally-contextual nonlinear CRFs for sequence labeling. Our approach directly incorporates information from the neighboring embeddings when predicting the label for a given word, and parametrizes the potential functions using deep neural networks. Our model serves as a drop-in replacement for the linear chain CRF, consistently outperforming it in our ablation study. On a variety of tasks, our results are competitive with those of the best published methods. In particular, we outperform the previous state of the art on chunking on CoNLL 2000 and named entity recognition on OntoNotes 5.0 English.

Compared to the CRF, the \CRFXO{} makes it easier to use locally-contextual information when predicting the label for a given word in a sentence. However if contextual embedding models such as BERT and Flair were jointly trained with sequence labeling tasks, this may eliminate the need for the additional locally-contextual connections of the \CRFXO{}; investigating this would be a particularly interesting avenue for future work.

The example sentences discussed in Section \ref{sec:abl} showed that the \CRFXO{} model could benefit from being combined with a source of external knowledge. Therefore another fruitful direction for future work could be to augment the \CRFXO{} with a knowledge base. This has been shown to improve performance on named entity recognition in particular \citep{WikiNER_Kazama_2007, KnowNER_Seyler_2017, KnowNER_He_2020}.

\bibliography{refs}
\bibliographystyle{icml2021}

\clearpage

\appendix

\section{Computational efficiency} \label{sec:apx:abl:comp}

\begin{table*}[t]
    \centering
    \begin{tabular}{l c@{\hskip 7pt}c@{\hskip 9pt}c c@{\hskip 7pt}c@{\hskip 9pt}c c@{\hskip 7pt}c@{\hskip 9pt}c c@{\hskip 7pt}c@{\hskip 9pt}c}
        \toprule
        \textsc{Model} & \multicolumn{3}{c}{\textsc{CoNLL 2000}} & \multicolumn{3}{c}{\textsc{Penn Treebank}} & \multicolumn{3}{c}{\textsc{CoNLL 2003}} & \multicolumn{3}{c}{\textsc{OntoNotes 5.0}} \\
         & \textsc{Prms} & \textsc{Trn} & \textsc{Inf} & \textsc{Prms} & \textsc{Trn} & \textsc{Inf} & \textsc{Prms} & \textsc{Trn} & \textsc{Inf} & \textsc{Prms} & \textsc{Trn} & \textsc{Inf} \\
        \midrule
        CRF(GloVe)              & $<$0.1 & 0.05 & 0.08 & $<$0.1 & 0.07 & 0.10 & $<$0.1 & 0.07 & 0.11 & $<$0.1 & 0.10 & 0.13 \\
        \CRFX{}(GloVe)          & $<$0.1 & 0.07 & 0.10 & $<$0.1 & 0.10 & 0.14 & $<$0.1 & 0.11 & 0.16 & $<$0.1 & 0.14 & 0.18 \\
        \CRFO{}(GloVe)          & 0.7    & 0.05 & 0.09 & 0.8    & 0.08 & 0.13 & 0.7    & 0.09 & 0.14 & 0.8    & 0.10 & 0.15 \\
        \CRFXO{}(GloVe)         & 2.2    & 0.10 & 0.15 & 2.3    & 0.13 & 0.18 & 2.2    & 0.16 & 0.20 & 2.3    & 0.17 & 0.22 \\[4pt]
        CRF(GloVe, biLSTM)      & 1.6    & 0.07 & 0.10 & 1.6    & 0.09 & 0.13 & 1.6    & 0.11 & 0.16 & 1.6    & 0.13 & 0.17 \\
        \CRFX{}(GloVe, biLSTM)  & 1.6    & 0.11 & 0.15 & 1.7    & 0.13 & 0.18 & 1.6    & 0.14 & 0.19 & 1.7    & 0.16 & 0.22 \\
        \CRFO{}(GloVe, biLSTM)  & 2.4    & 0.07 & 0.12 & 2.4    & 0.11 & 0.16 & 2.4    & 0.13 & 0.17 & 2.4    & 0.13 & 0.18 \\
        \CRFXO{}(GloVe, biLSTM) & 3.8    & 0.12 & 0.18 & 3.9    & 0.16 & 0.22 & 3.8    & 0.16 & 0.23 & 3.9    & 0.20 & 0.28 \\[4pt]
        CRF(BERT)               & $<$0.1 & 0.05 & 0.09 & $<$0.1 & 0.12 & 0.17 & $<$0.1 & 0.13 & 0.19 & $<$0.1 & 0.18 & 0.23 \\
        \CRFX{}(BERT)           & 0.1    & 0.09 & 0.14 & 0.1    & 0.14 & 0.21 & $<$0.1 & 0.16 & 0.23 & 0.1    & 0.18 & 0.26 \\
        \CRFO{}(BERT)           & 1.3    & 0.07 & 0.11 & 1.3    & 0.15 & 0.20 & 1.3    & 0.15 & 0.21 & 1.3    & 0.18 & 0.25 \\
        \CRFXO{}(BERT)          & 3.9    & 0.11 & 0.18 & 4.0    & 0.22 & 0.29 & 3.9    & 0.22 & 0.31 & 4.0    & 0.23 & 0.33 \\[4pt]
        CRF(Flair)              & 0.1    & 0.20 & 0.27 & 0.2    & 0.36 & 0.42 & $<$0.1 & 0.46 & 0.58 & 0.2    & 0.50 & 0.64 \\
        \CRFX{}(Flair)          & 0.3    & 0.29 & 0.39 & 0.6    & 0.38 & 0.43 & 0.1    & 0.55 & 0.65 & 0.5    & 0.56 & 0.67 \\
        \CRFO{}(Flair)          & 5.5    & 0.25 & 0.35 & 5.6    & 0.36 & 0.45 & 5.4    & 0.46 & 0.60 & 5.6    & 0.54 & 0.66 \\
        \CRFXO{}(Flair)         & 16.5   & 0.35 & 0.48 & 16.9   & 0.52 & 0.63 & 16.3   & 0.62 & 0.74 & 16.7   & 0.68 & 0.81 \\
        \bottomrule
    \end{tabular}
    \caption{Computational efficiency statistics of each the models trained. The \textsc{Prms} column shows the number of parameters, in millions. \textsc{Trn} shows the average time taken, in seconds, for a training iteration with a batch of 128 sentences. \textsc{Inf} shows the average time taken, in seconds, to infer the most likely labels for a batch of 256 sentences. All values were computed using the same GPU.}
    \label{tab:apx:abl:comp}
\end{table*}

Table \ref{tab:apx:abl:comp} shows the detailed computational efficiency statistics of each of the models we train. For a given embedding model, we find that the \CRFXO{} takes approximately 1.5 to 2 times longer for each training and inference iteration than the CRF. When not using the bidirectional LSTM, the \CRFX{} and \CRFXO{} have approximately 3 times as many parameters as the CRF and \CRFO{} respectively. However when using the bidirectional LSTM, this component dominates the number of parameters. This means that the \CRFX{}(GloVe, biLSTM) has approximately the same number of parameters as the CRF(GloVe, biLSTM) and the \CRFXO{}(GloVe, biLSTM) only has approximately 1.6 times as many parameters as the \CRFO{}(GloVe, biLSTM).

\section{A more general parametrization} \label{sec:apx:alt:param}

\begin{figure}[t]
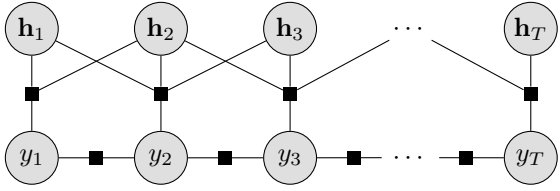

    \centering
    \tikz{%
        \node[obs] (h1) {$\h_{1}$};
        \node[obs, right=of h1] (h2) {$\h_{2}$};
        \node[obs, right=of h2] (h3) {$\h_{3}$};
        \node[const, right=of h3] (hdots) {$\cdots$};
        \node[obs, right=of hdots] (hT) {$\h_{T}$};
        \node[obs, below=of h1] (y1) {$y_{1}$};
        \node[obs, below=of h2] (y2) {$y_{2}$};
        \node[obs, below=of h3] (y3) {$y_{3}$};
        \node[const, below=of hdots, yshift=-15.5pt] (ydots) {$\cdots$};
        \node[obs, below=of hT] (yT) {$y_{T}$};
        \factor[above=of y1] {y1h} {} {} {};
        \factor[above=of y2] {y2h} {} {} {};
        \factor[above=of y3] {y3h} {} {} {};
        \factor[above=of yT] {yTh} {} {} {};
        \edge[-] {h1} {y1h};
        \edge[-] {h2} {y1h};
        \edge[-] {y1h} {y1};
        \edge[-] {h1} {y2h};
        \edge[-] {h2} {y2h};
        \edge[-] {h3} {y2h};
        \edge[-] {y2h} {y2};
        \edge[-] {h2} {y3h};
        \edge[-] {h3} {y3h};
        \edge[-, shorten <= 4pt] {hdots} {y3h};
        \edge[-] {y3h} {y3};
        \edge[-, shorten <= 4pt] {hdots} {yTh};
        \edge[-] {hT} {yTh};
        \edge[-] {yTh} {yT};
        \factor[right=of y1] {y1y} {} {} {};
        \factor[right=of y2] {y2y} {} {} {};
        \factor[right=of y3] {y3y} {} {} {};
        \factor[right=of ydots] {ydotsy} {} {} {};
        \edge[-] {y1} {y2};
        \edge[-] {y2} {y3};
        \edge[-, shorten >= 4pt] {y3} {ydots};
        \edge[-, shorten <= 4pt] {ydots} {yT};
    }
    \caption{The graphical model of the \CRFXO{}-Concat.}
    \label{fig:apx:alt:param}
\end{figure}

\begin{table*}[t]
    \centering
    \begin{tabular}{lcccc}
        \toprule
        \textsc{Model} & \textsc{CoNLL 2000} & \textsc{Penn Treebank} & \textsc{CoNLL 2003} & \textsc{OntoNotes 5.0} \\
         & \textsc{F1} & \textsc{Accuracy} & \textsc{F1} & \textsc{F1} \\
        \midrule
        \CRFXO{}(BERT)         & 97.40 & 97.24 & 93.82 & 92.17 \\
        \CRFXO{}-Concat(BERT)  & 97.23 & 97.10 & 93.10 & 91.80 \\[4pt]
        \CRFXO{}(Flair)        & 97.52 & 97.56 & 94.22 & 91.54 \\
        \CRFXO{}-Concat(Flair) & 97.48 & 97.44 & 93.90 & 91.30 \\
        \bottomrule
    \end{tabular}
    \caption{Test set results with a more general parametrization of the local context.}
    \label{tab:apx:alt:param}
\end{table*}

A more general parametrization of the local context would, at each time step, have a single potential for $\h_{t-1}$, $\h_{t}$, $\h_{t+1}$ and $y_{t}$ instead of three separate ones as in Section \ref{sec:model}. 

The graph of this model, which we refer to as \CRFXO{}-Concat, is shown in Figure \ref{fig:apx:alt:param}. We define the following potential: \begin{align}
    \sigmat = \sigmatth
\end{align}

Then, the model is parametrized as follows: \begin{align}
    p(\y|\x,\theta) &= \frac{\prod_{t} \psit \sigmat}{\sum_{\y} \prod_{t} \psit \sigmat}
\end{align}

The parametrization of $\log \psit$ remains the same as in Equation (\ref{eq:crf:psi}). We parametrize $\log \sigmat$ as follows: \begin{align}
    \log \sigmat = \ii(y_{t})^{\mathsf{T}} \fsigma([\h_{t-1}; \h_{t}; \h_{t+1}])
\end{align}

$\fsigma$ is a feedforward network which takes as input the concatenated embeddings $\h_{t-1}$, $\h_{t}$, and $\h_{t+1}$.

We train this model with BERT and Flair embeddings. As per Section \ref{sec:exp:model}, $\fsigma$ has 2 layers with 600 units, ReLU activations and a skip connection. This means that for any choice of word embeddings, \CRFXO{} and \CRFXO{}-Concat have approximately the same number of parameters.

The results are shown in Table \ref{tab:apx:alt:param}. We see that in all cases, \CRFXO{} performs better than \CRFXO{}-Concat. This result is likely due to \CRFXO{} being easier to train than \CRFXO{}-Concat: as well as having better test set scores, we find that \CRFXO{} consistently achieves a higher value for the training objective than \CRFXO{}-Concat, which suggests that it finds a better local optimum.

\section{Widening the contextual window} \label{sec:apx:alt:wide}

\begin{figure}[t]
    \centering
    \tikz{%
        \node[obs] (h1) {$\h_{1}$};
        \node[obs, right=of h1] (h2) {$\h_{2}$};
        \node[obs, right=of h2] (h3) {$\h_{3}$};
        \node[const, right=of h3] (hdots) {$\cdots$};
        \node[obs, right=of hdots] (hT) {$\h_{T}$};
        \node[obs, below=of h1] (y1) {$y_{1}$};
        \node[obs, below=of h2] (y2) {$y_{2}$};
        \node[obs, below=of h3] (y3) {$y_{3}$};
        \node[const, below=of hdots, yshift=-15.5pt] (ydots) {$\cdots$};
        \node[obs, below=of hT] (yT) {$y_{T}$};
        \edge[-] {h1} {y1};
        \edge[-] {h2} {y2};
        \edge[-] {h3} {y3};
        \edge[-] {hT} {yT};
        \edge[-] {h1} {y2};
        \edge[-] {h2} {y3};
        \edge[-, shorten >= 4pt] {h3} {ydots};
        \edge[-, shorten <= 4pt] {hdots} {yT};
        \edge[-] {y1} {h2};
        \edge[-] {y2} {h3};
        \edge[-, shorten >= 4pt] {y3} {hdots};
        \edge[-, shorten <= 4pt] {ydots} {hT};
        \edge[-] {h1} {y3};
        \edge[-, shorten >= 4pt] {h2} {ydots};
        \edge[-] {y1} {h3};
        \edge[-, shorten >= 4pt] {y2} {hdots};
        \edge[-] {y1} {y2};
        \edge[-] {y2} {y3};
        \edge[-, shorten >= 4pt] {y3} {ydots};
        \edge[-, shorten <= 4pt] {ydots} {yT};
    }
    \caption{The graphical model of the \CRFXO{}-Wide.}
    \label{fig:apx:alt:wide}
\end{figure}

\begin{table*}[t]
    \centering
    \begin{tabular}{lcccc}
        \toprule
        \textsc{Model} & \textsc{CoNLL 2000} & \textsc{Penn Treebank} & \textsc{CoNLL 2003} & \textsc{OntoNotes 5.0} \\
         & \textsc{F1} & \textsc{Accuracy} & \textsc{F1} & \textsc{F1} \\
        \midrule
        \CRFXO{}(BERT)       & 97.40 & 97.24 & 93.82 & 92.17 \\
        \CRFXO{}-Wide(BERT)  & 97.39 & 97.19 & 93.81 & 92.15 \\[4pt]
        \CRFXO{}(Flair)      & 97.52 & 97.56 & 94.22 & 91.54 \\
        \CRFXO{}-Wide(Flair) & 97.50 & 97.53 & 94.23 & 91.44 \\
        \bottomrule
    \end{tabular}
    \caption{Test set results with a wider local context.}
    \label{tab:apx:alt:wide}
\end{table*}

The \CRFXO{} as described in Section \ref{sec:model} has a local context of width 3. That is, the embeddings $\h_{t-1}$, $\h_{t}$ and $\h_{t+1}$ are directly used when modeling the label $y_{t}$. One may consider that the performance of the model would be better with a wider context. 

The graph of this model, which we refer to as \CRFXO{}-Wide, is shown in Figure \ref{fig:apx:alt:wide}. We define the following potentials, in addition to those in Equations (\ref{eq:crf:def_psi}), (\ref{eq:crf:def_eta}), (\ref{eq:model:def_phi}) and (\ref{eq:model:def_xi}): \begin{align}
    \pit &= \pitth \\
    \zetat &= \zetatth
\end{align}

The model is then parametrized as: \begin{align}
    p(\y|\x,\theta) &= \frac{\prod_{t} \psit \pit \phit \etat \xit \zetat}{\sum_{\y} \prod_{t} \psit \pit \phit \etat \xit \zetat}
\end{align}

As with the parametrization of the potentials $\phit$, $\etat$ and $\xit$ in Equations (\ref{eq:model:phi}) to (\ref{eq:model:xi}), $\log \pit$ and $\log \zetat$ are parametrized using feedforward networks $\fpi$ and $\fzeta$ whose inputs are the embeddings $\h_{t-2}$ and $\h_{t+2}$ respectively: \begin{align}
    \log \pit = \ii(y_{t})^{\mathsf{T}} \fpi(\h_{t-2}) \\
    \log \zetat = \ii(y_{t})^{\mathsf{T}} \fzeta(\h_{t+2})
\end{align}

We train this model with BERT and Flair embeddings. As per Section \ref{sec:exp:model}, $\fpi$ and $\fzeta$ each have 2 layers with 600 units, ReLU activations and a skip connection.

The results are shown in Table \ref{tab:apx:alt:wide}. In general, we see that there is very little difference in performance between \CRFXO{} and \CRFXO{}-Wide. This suggests that the local context used by the \CRFXO{} is sufficiently wide for the tasks evaluated.

\end{document}